\documentclass[10pt,twocolumn,letterpaper]{article}

\usepackage{cvpr}              
\usepackage{multirow}
\usepackage{multicol}
\usepackage[accsupp]{axessibility}

\definecolor{cvprblue}{rgb}{0.21,0.49,0.74}
\usepackage[pagebackref,breaklinks,colorlinks,allcolors=cvprblue]{hyperref}

\def\paperID{12612} 
\def\confName{CVPR}
\def\confYear{2025}

\title{MVPortrait: Text-Guided Motion and Emotion Control for\\ Multi-view Vivid Portrait Animation}

\author{
  Yukang Lin$^{1,}$\thanks{Equal contribution.}, 
  Hokit Fung$^{2,}$\footnotemark[1], 
  Jianjin Xu$^{3}$, 
  Zeping Ren$^{1}$, 
  Adela S.M. Lau$^{2}$, 
  Guosheng Yin$^{2}$\thanks{Corresponding authors.}, 
  Xiu Li$^{1}$\footnotemark[2] 
  \\  
  $^{1}$Tsinghua University, 
  $^{2}$The University of Hong Kong, 
  $^{3}$Carnegie Mellon University\\
  {\tt\small \{linyk23,rzp22\}@mails.tsinghua.edu.cn,}
  {\tt\small fung0311@connect.hku.hk, jianjinx@andrew.cmu.edu, } \\
  {\tt\small \{adelalau, gyin\}@hku.hk, }
  {\tt\small li.xiu@sz.tsinghua.edu.cn}
}

\begin{document}
\maketitle
\begin{abstract}
Recent portrait animation methods have made significant strides in generating realistic lip synchronization. However, they often lack explicit control over head movements and facial expressions, and cannot produce videos from multiple viewpoints, resulting in less controllable and expressive animations. Moreover, text-guided portrait animation remains underexplored, despite its user-friendly nature. We present a novel two-stage text-guided framework, MVPortrait (Multi-view Vivid Portrait), to generate expressive multi-view portrait animations that faithfully capture the described motion and emotion. MVPortrait is the first to introduce FLAME as an intermediate representation, effectively embedding facial movements, expressions, and view transformations within its parameter space. In the first stage, we separately train the FLAME motion and emotion diffusion models based on text input. In the second stage, we train a multi-view video generation model conditioned on a reference portrait image and multi-view FLAME rendering sequences from the first stage. Experimental results exhibit that MVPortrait outperforms existing methods in terms of motion and emotion control, as well as view consistency. Furthermore, by leveraging FLAME as a bridge, MVPortrait becomes the first controllable portrait animation framework that is compatible with text, speech, and video as driving signals.
\end{abstract}    
\section{Introduction}
\label{sec:intro}

With the advent of high-quality datasets \cite{shao2024human4dit, li2023finedance} and advanced methods \cite{hu2024animate, li2024lodge, jin2024alignment, xu2024mambatalk, li2024lodge++, xu2024chain}, digital human animation has achieved promising results. Among animations, portrait animation intends to generate lively video from a reference image. With different guidance such as audio, textual instructions, and exemplar gestures, the static portrait is endowed with speaking and moving capabilities. 


\begin{figure}[t]
    \centering
    \includegraphics[width=\linewidth]{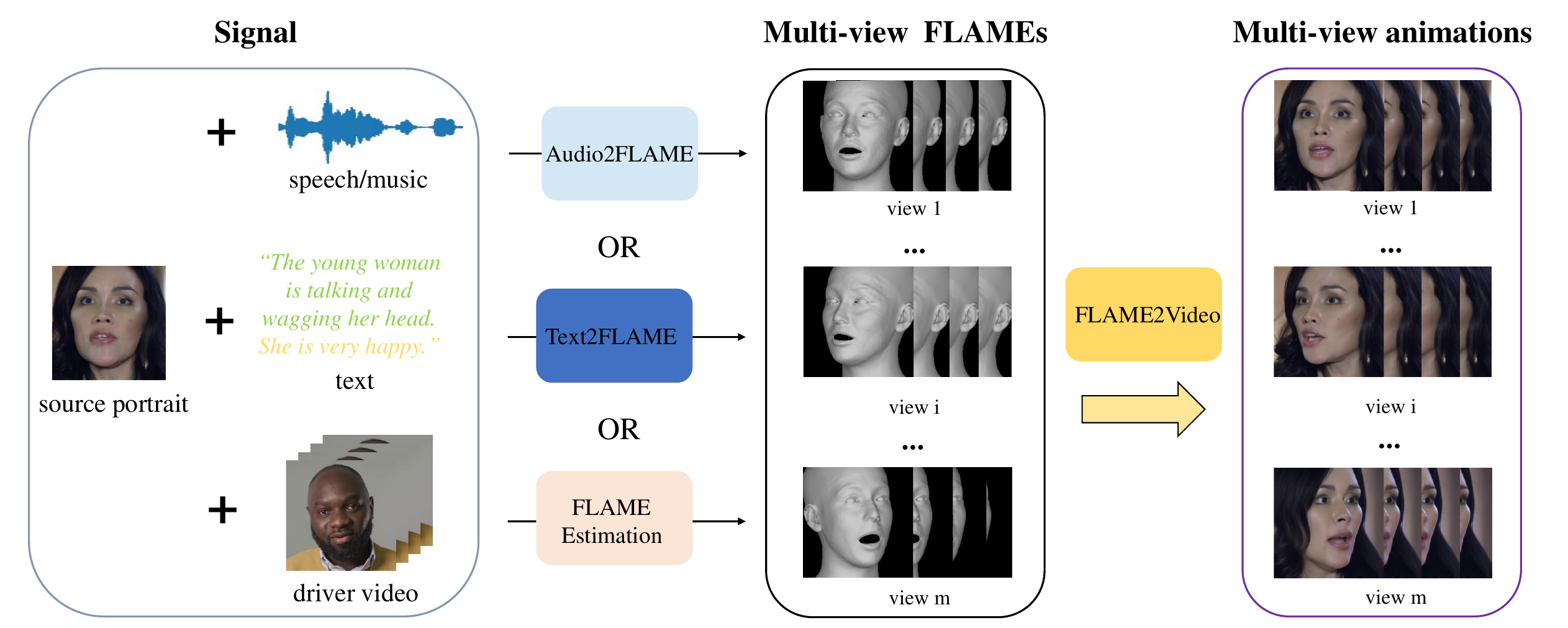}
    \caption{The unified pipeline of multi-view portrait animation. Users can obtain FLAME sequences via audio-driven methods like Talkshow \cite{yi2023generating} or FLAME estimation methods like DECA \cite{deca} from driver video. This paper focuses on text-driven animation.}
    \label{fig:unify}
\end{figure}

There are three primary approaches to accomplish portrait animation: (1) \textit{Emotion control especially lip synchronization} \cite{xu2024hallo, xu2024vasa, wang2024v}, (2) \textit{Motion control} \cite{wei2024aniportrait, ma2024follow}, and (3) \textit{3D-aware face reconstruction} \cite{deng2024portrait4d, bhattarai2024triplanenet}. The first approach focuses on aligning lip movements with audio input. The common way is to extract the audio features and inject the features into the generative network, often via a cross-attention layer in UNet \cite{ronneberger2015u}. Compared to full facial performance, lip synchronization is a restrictive subset and lacks the generalization of portrait expression learning. The second approach relies on pose images, i.e. the expression-aware landmarks, to direct the animation, producing lifelike facial expressions and diverse postures. However, relying on landmarks alone may not capture all facial details, potentially limiting visual fidelity. The third approach generates animations by creating sequential 3D representations like Triplane \cite{shue20233d} and obtains multi-view renderings.
However, Triplane can introduce inconsistent artifacts, such as the ``multi-face'' issue in side views \cite{an2023panohead}, which may degrade performance in multi-view animations. It is worth noting that text-guided portrait animation, as the most user-friendly and flexible scheme, has not been fully explored.

To address the problems of restrictive facial expression, lack of detailed guidance, and poor multi-view performance, we propose \textbf{MVPortrait}, a novel two-stage method designed to control facial motion and emotion through text guidance and generate vivid portrait animation from different viewpoints. It is important to note that MVPortrait utilizes FLAME \cite{FLAME} as an intermediate representation. FLAME is a 3D parametric model that compactly represents face shape, pose, and expression, all unified within its parameter space. Benefiting from FLAME's extensive use in speech-driven head generation and facial reconstruction from images, our framework also supports audio-driven and video-driven portrait animation. 
The unified framework for text-driven, audio-driven, and video-driven is shown in \cref{fig:unify}. Furthermore, the FLAME representation allows for rendering multi-view images by simply adjusting the head pose in the yaw, roll, and pitch directions, which makes it highly convenient for our multi-view animation task.

Our MVPortrait consists of two stages: Text2FLAME and FLAME2Video. In the Text2FLAME stage, in contrast to prior work which encodes the text instruction into hidden features, we separately train two diffusion models in the FLAME parameter space, namely MotionDM and EmotionDM. MotionDM takes the motion descriptions as input and outputs a sequence of pose parameters, while EmotionDM produces a sequence of expression parameters from emotion descriptions. These two models focus on capturing the full range of head movement and expression dynamics and generate a FLAME sequence that aligns with the text prompt. In the FLAME2Video stage, we first obtain multi-view FLAME conditions by rendering the FLAME sequence from multiple viewpoints. Subsequently, we proceed with training the multi-view animation framework, which includes the VAE, FLAME encoder, Reference UNet, and Denoising UNet. By utilizing the FLAME encoder, the pose and expression information captured in the FLAME rendering can be effectively injected into the Denoising UNet. To ensure consistency in multi-view animations, we introduce the view attention mechanism by incorporating a view module within the Denoising UNet. 



Our contributions can be summarized as follows:


\begin{itemize}
\item Through the implementation of MotionDM and EmotionDM, we integrate facial motion and expression control into our unified text-guided framework, thereby expanding the controllability for avatar manipulation.


\item We extend the current portrait animation to the multi-view level, enhancing its versatility and practicality. This expansion enables a richer, more immersive visual experience by representing subjects from multiple perspectives.

\item We introduce FLAME as an intermediate in our two-stage framework, allowing MVPortrait to accommodate various forms of signals, including audio and driver video, as illustrated in \cref{fig:unify}. This integration highlights enhanced adaptability compared to previous methodologies.


\end{itemize}

\section{Related Work}
\subsection{Text-conditioned Human Motion Generation}

Recently, significant advancements have been made in text-conditioned human motion generation. Earlier works focused on training models using motion capture datasets~\cite{ionescu2013human3, de2009guide} without text labels, treating the problem as a deterministic mapping task utilizing CNN-based~\cite{holden2015learning, holden2016deep} or RNN-based~\cite{ghosh2017learning, li2017auto} models. 
The release of text-labeled motion datasets, such as HumanML3D~\cite{guo2022generating} and KIT-ML~\cite{plappert2016kit}, has accelerated development in this area~\cite{chen2023executing, zhang2023generating, wang2023fg, zhong2023attt2m}. Notably, MotionDiffuse~\cite{zhang2022motiondiffuse} and MDM~\cite{tevet2023human} were the first to apply diffusion models to the text-to-motion task. ReMoDiffuse~\cite{zhang2023remodiffuse} enhanced this approach by retrieving motions that correspond to target text and generating motion based on those features, while CrossDiff~\cite{ren2023realistic} utilized 2D motion data to support 3D motion generation. 
However, there has been no dataset or significant work dedicated to text-driven facial motion generation. We aim to leverage the successes in motion generation to fill this gap.

\begin{figure*}[h]
    \includegraphics[width=\textwidth]{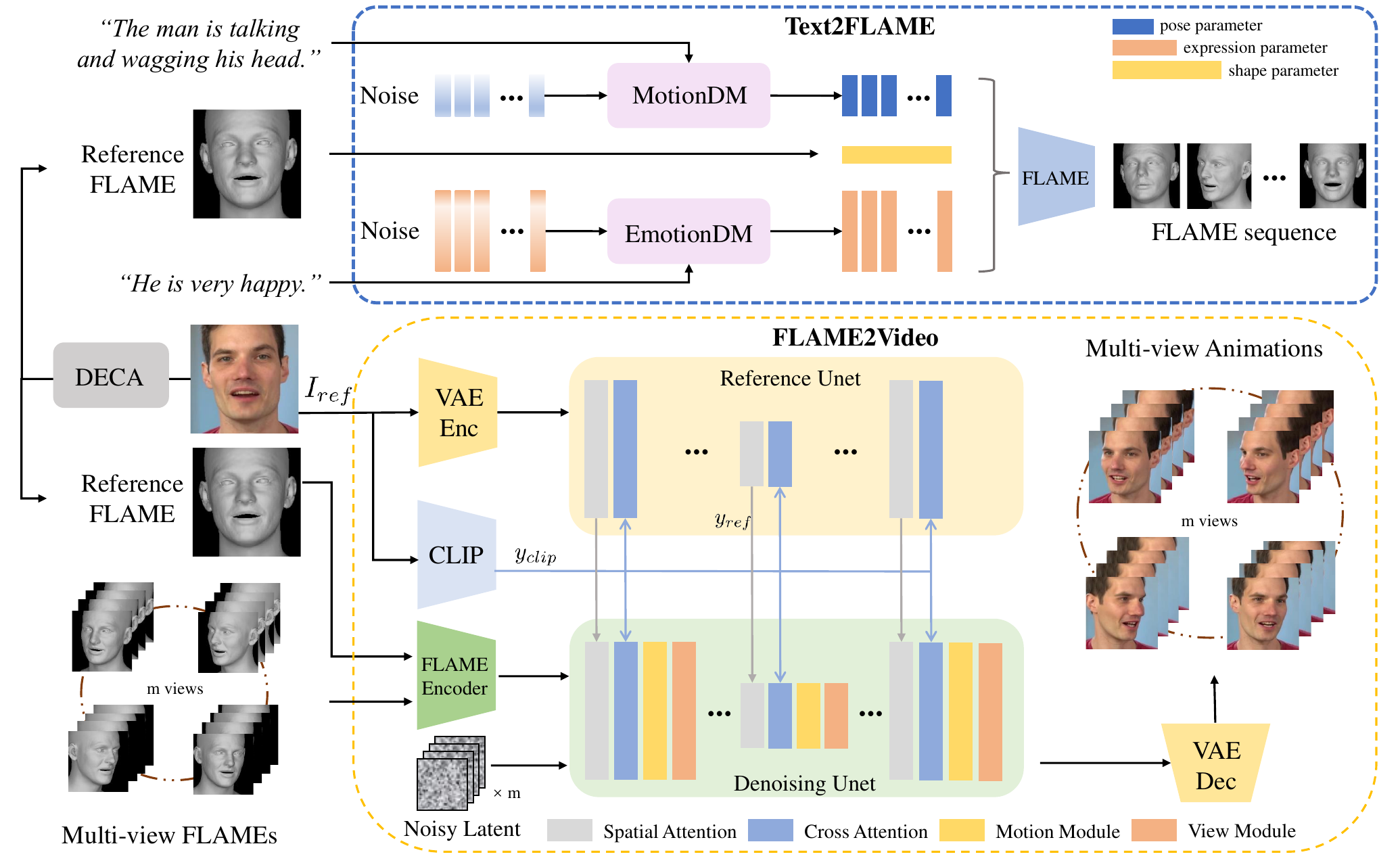}
    \caption{\textbf{The overview of MVPortrait.} MVPortrait consists of two stages: Text2FLAME and FLAME2Video. In the Text2FLAME stage, a reference FLAME is first estimated from the reference image. The text prompt is divided into motion and emotion descriptions, which are then used by MotionDM and EmotionDM to generate the corresponding pose and expression sequences. These sequences, combined with the reference FLAME's shape, form the FLAME sequence. In the FLAME2Video stage, the reference image, aligned reference FLAME rendering, and multi-view renderings of the FLAME sequence are used as inputs to generate multi-view vivid and consistent animations.}
    \label{pipeline}
\end{figure*}

\subsection{Portrait Animation}

The great advancement of image generation has also extended to the realm of video generation \cite{blattmann2023align, guo2023animatediff, blattmann2023stable, wang2024taming, xu2025hunyuanportrait, wang2024cove} by integrating the Transformer architecture \cite{vaswani2017attention} or the VAEs framework \cite{kingma2013auto}. There are typically three branches for portrait image animation: (1) Emotion control especially lip synchronization, (2) Motion control, and (3) 3D-aware face reconstruction. 

The first branch refers to talking head generation. LipSyncExpert \cite{prajwal2020lip,suwajanakorn2017synthesizing} introduced the lip-synchronization network to align the talking audio to the mouth movement. 
Researchers \cite{zhang2023sadtalker, zhou2021pose, yu2023talking} later attempt to model different parts of the face. VASA-1 \cite{xu2024vasa} and EMO \cite{tian2024emo} utilize robust transformer models and cross-attention mechanisms to capture various facial components within the motion latent space. Motion control has been further developed on the foundation of the talking head branch, emphasizing enhancing spatial and temporal consistency. FollowYourEmoji \cite{ma2024follow} and AniPortrait \cite{wei2024aniportrait} have incorporated motion and expression-aware condition maps to allow for personalized character motion adjustments. The advancement of 3D-aware face reconstruction techniques allows 2D portrait animation to aggregate the domains of 3D reconstruction \cite{deng2024portrait4d, bhattarai2024triplanenet, lin2024consistent123, lin2023rich}. Rather than concentrating on creating intricate 3D head models, these approaches utilize expedited methods such as 3DMM \cite{blanz2003face} and 3D GAN \cite{chan2022efficient} that project 3D facial features. SadTalker \cite{zhang2023sadtalker} was the early trial to use lip-only 3DMM coefficients.  The 3D facial mesh was used in AniPortrait \cite{wei2024aniportrait} to predict the reference and target pose for better expression alignments. Portrait4D-v2 \cite{deng2024portrait4d} turns the sequential portrait animation into temporal 3D face Triplane \cite{shue20233d} and renders the views given the camera input.

\section{Methodology}
Given a reference image $I_{\rm ref}$ and text prompt $y$ describing motion and emotion, our goal is to synthesize vivid multi-view videos $V$ with the appearance of $I_{\rm ref}$ while adhering to $y$ and maintaining multi-view consistency. As illustrated in \cref{pipeline}, the proposed framework comprises two modules, namely Text2FLAME and FLAME2Video. 


\subsection{FLAME Representation}
This study utilizes FLAME as an intermediate representation to synthesize portrait animation based on rendering images. FLAME is a statistical 3D head model that integrates distinct linear spaces for identity shape and expression, using linear blend skinning and pose-dependent corrective blend shapes to animate the neck and jaw. Given parameters of facial shape $\boldsymbol{\beta} \in \mathbb{R}^{|\boldsymbol{\beta}|}$, pose $\boldsymbol{\theta}\in\mathbb{R}^{3k+3}$ (with $k=4$ joints for neck, jaw, and eyeballs), and expression $\boldsymbol{\psi}\in\mathbb{R}^{|\boldsymbol{\psi}|}$, FLAME outputs a mesh with $n = 5023$ vertices. The model is defined as
\begin{equation}
M(\boldsymbol{\beta},\boldsymbol{\theta},\boldsymbol{\psi}) = W(T_P(\boldsymbol{\beta},\boldsymbol{\theta},\boldsymbol{\psi}), \mathbf{J}(\boldsymbol{\beta}), \boldsymbol{\theta}, \mathcal{W}),
\end{equation}

\begin{small}
\begin{equation}
T_P(\boldsymbol{\beta},\boldsymbol{\theta},\boldsymbol{\psi}) = T + B_S(\boldsymbol{\beta};\boldsymbol{S}) + B_P(\boldsymbol{\theta};\boldsymbol{P}) + B_E(\boldsymbol{\psi};\boldsymbol{E}).
\end{equation}
\end{small}
The blend skinning function \(W(T_P,\mathbf{J}, \boldsymbol{\theta},\mathcal{W})\) rotates the vertices in \(T_P\in\mathbb{R}^{3n}\) around joints \(\mathbf{J}\in\mathbb{R}^{3k}\), linearly smoothed by blendweights \(\mathcal{W}\in\mathbb{R}^{k\times n}\). The mean template is denoted as \(T \in \mathbb{R}^{3n}\), while the shape, pose, and expression blend shapes are represented by \(B_S(\boldsymbol{\beta};\boldsymbol{S})\), \(B_P(\boldsymbol{\theta};\boldsymbol{P})\), and \(B_E(\boldsymbol{\psi};\boldsymbol{E})\), respectively. 

In our experiment, we leverage the pose parameter $f_{\rm pose} = \boldsymbol{\theta}$ for head motion control, expression parameter $f_{\rm exp} = \boldsymbol{\psi}$ for emotion control, and shape parameter $f_{\rm shape} = \boldsymbol{\beta}$ for facial identity. This enables the transformation of motion and expression control into the generation of FLAME pose and expression parameters, which are then rendered into images, explicitly capturing facial movements and expressions for vivid and controllable animations.

\subsection{Text2FLAME}
In this part, we aim to synthesize FLAME sequence $f^{1:N} = \{f_{\rm shape}^{1:N}\, f_{\rm pose}^{1:N}, f_{\rm exp}^{1:N}\} $ of length $N$ given text condition $y$. For facial shape, we estimate $f_{\rm shape}$ from reference portrait image $I_{\rm ref}$ using DECA \cite{deca}, and utilize it for FLAME blending. Due to the varying degrees of motion amplitude and emotion intensity, we decouple the generation of pose $f_{\rm pose}$ and expression $f_{\rm exp}$. 

\begin{figure}[h]
    \centering
    \includegraphics[width=\linewidth]{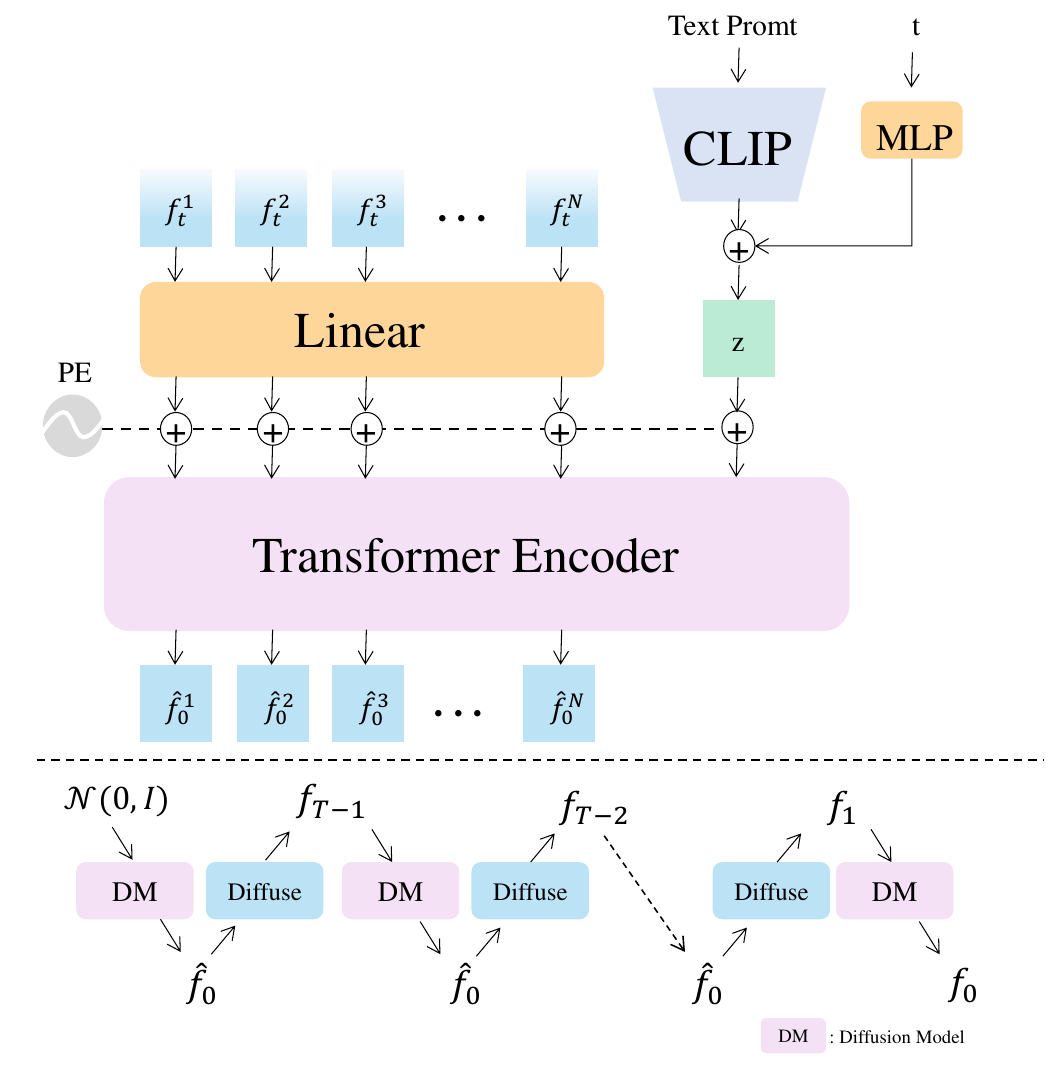}
    \caption{\textbf{(Top) The framework of $DM$.} MotionDM and EmotionDM both employ MDM as the backbone. They denoise conditioned on motion or emotion descriptions separately. \textbf{(Bottom) Sampling process.} Given a condition, $DM$ denoises $f_T$ from a Gaussian distribution to obtain the clean motion or emotion $f_0$.}
    \label{fig: mdm}
\end{figure}

To convert text into pose and expression, we adopt the transformer-based diffusion architecture \cite{tevet2023human}, which not only excels in the quality and diversity of results but is also remarkably lightweight and controllable.
It is worth noting that to decouple motion and emotion, we train two separate diffusion models, namely MotionDM and EmotionDM, which share the same network architecture. The generated $f_{\rm type}^{1:N}=\{f^i\}_{i=1}^N$ is a sequence of facial poses or expressions represented by $f_{\rm type}^i\in \mathbb{R}^{D}$, where type $=$ \{pose, exp\}, $f_{\rm pose}^i \in \mathbb{R}^{12}$ and $f_{\rm exp}^i \in \mathbb{R}^{50}$. For simplicity, we omit the subscript ``type'' in the following sections and refer to both MotionDM and EmotionDM as $DM$. The sampling strategy and architecture of $DM$ are shown in \cref{fig: mdm}. 

Based on the DDPM \cite{ho2020denoising} framework, $DM$ assumes $T$ diffusion steps modeled by the Markov noising process,
\begin{equation}
    q(f_t^{1:N}|f_{t-1}^{1:N})=\mathcal{N}(\sqrt{\alpha_t}f_{t-1}^{1:N},(1-\alpha_t){\bf I}),
\end{equation}
where $\alpha_t\in(0,1)$ are constant hyper-parameters from the noise schedule. From here on, we use $f_t$ to denote the complete sequence at noising step $t$. In our task, $DM$ models the distribution $p(f_0|y)$ as the reversed process of gradually cleaning $f_T$. At each denoising step, we obtain $\hat{f}_0=DM(f_t,t,y)$ with the simple loss,
\begin{equation} \mathcal{L}_{\mathrm{simple}}=E_{f_0\sim q(f_0|y),t\sim[1,T]}[\|f_0-DM(f_t,t,y)\|_2^2].
\end{equation}
To mitigate jittery and discontinuity, we employ a velocity geometric loss $\mathcal{L}_\text{vel}$ to regularize $DM$,
\begin{equation}
    \mathcal{L}_\text{vel}=\frac1{N-1}\sum_{i=1}^{N-1}\|(f_0^{i+1}-f_0^i)-(\hat{f}_0^{i+1}-\hat{f}_0^i)\|_2^2.
\end{equation}
Given the tuning parameter $\lambda_{\mathrm{vel}}$, the overall training loss of $DM$ is
\begin{equation}
\mathcal{L}_{DM}=\mathcal{L}_{\mathrm{simple}}+\lambda_{\mathrm{vel}}\mathcal{L}_{\mathrm{vel}}.
\end{equation}

The experiments reveal that noise in raw data can cause severe jittery and flickering in results, prompting the use of a sliding window approach for smoothing. Given a window size $L$, the smoothed value $\tilde{x}_i$ at position $i$ is computed as the average of the original data values within the window centered at $i$. The formula for the smoothed value is
\begin{equation}
\tilde{f}_i = \frac{1}{L} \sum_{j=i - {L}/{2}}^{i + {L}/{2}} f_j.
\end{equation}

Once MotionDM and EmotionDM are trained, by inputting text $y$, we can sample FLAME motion and expression sequences, which together define the FLAME model's performance. According to the experiment, the generated FLAME sequences effectively reflect the text description $y$.

\subsection{FLAME2Video}
Given multi-view FLAME rendering sequences $P = [ p_1^{1:N}, \ldots, p_m^{1:N}]$, our objective is to generate temporally consistent and multi-view consistent portrait animations $V = [I_1^{1:N}, \ldots, I_m^{1:N}]$ with $m$ representing the number of views, while the appearance in the videos adheres to $I_{\text{ref}}$. To achieve the realistic animation, we follow the recent diffusion-based portrait animation approaches \cite{xu2024magicanimate, hu2024animate, ma2024follow, wei2024aniportrait}, and adopt the reference UNet, pose guider and motion module in our framework. Additionally, we insert the view module into the denoising UNet to ensure the consistency of different views.

\textbf{Appearance control.} To preserve the appearance information of $I_{\rm ref}$, such as face identity and background, reference UNet creates a trainable copy of denoising UNet to compute low-level features of $I_{\rm ref}$. Given the hidden states $z_t$ from each spatial self-attention layer of denoising UNet and reference feature $y_{\rm ref}$ from the corresponding layer of reference UNet, the appearance injection is mathematically formulated as
\vspace{-2mm}
\begin{equation}
\mathrm{Attention}(Q,K,V,y_{\rm ref}) = \mathrm{Softmax}\left(\frac{QK^{\prime \top}}{\sqrt{d}}\right)V^{\prime},
\end{equation}
\vspace{-3mm}
\begin{equation}
Q=W^{Q}z_{t},K^{\prime}=W^{K}[z_{t},y_{\rm ref}],V^{\prime}=W^{V}[z_{t},y_{\rm ref}].
\end{equation}
where $[\cdot]$ denotes concatenation operation. Moreover, we also utilize the high-level reference features $y_{\rm clip}$ extracted via CLIP \cite{radford2021learning} to adapt the cross-attention layers of both reference and denoising UNet.

\textbf{Pose and expression guidance.} We incorporate the rendering image from FLAME as the pose and expression driving signal. Obtaining the FLAME sequence $f^{1:N}$ from the Text2FLAME stage, we can generate the multi-view FLAME rendering sequences by $P = \mathbf{R}(f^{1:N};C)$, where $C = \{c_1, \ldots, c_m\}$ denotes the set of perspectives.
The architecture of our FLAME encoder is the same as the pose guider of AniPortrait \cite{wei2024aniportrait}. 
The feature map output from the FLAME encoder is added to the noisy latent and then passed into the denoising UNet. The FLAME encoder captures the facial shape, pose, and expression from the FLAME rendering, constraining the generated video to align with the reference FLAME sequences. This approach ensures that the video’s facial shape remains consistent with the reference image while precisely following the head movements and facial expressions described in the text condition $y$.

\textbf{Motion module.} We follow AnimateDiff \cite{guo2023animatediff} by inserting temporal layers into Stable Diffusion 1.5, transforming the FLAME sequence into a temporally consistent and photorealistic animated portrait. Now we focus solely on the single-view scenario. Given a feature map $x \in \mathbb{R}^{b\times t\times h\times w\times c}$ from denoising UNet, we reshape it to $x \in \mathbb{R}^{(b\times h\times w)\times t\times c}$ before the motion module, and then perform temporal attention, which refers to self-attention along the dimension $t$.

\textbf{View module.} From here on, we explain the generation of multi-view videos. Specifically, we insert a view module after the motion module, where the architecture of the view module is identical to that of the motion module. The multi-view feature map, initially $x \in \mathbb{R}^{b\times m\times t\times h\times w\times c}$, is reshaped to $x \in \mathbb{R}^{(b\times m)\times t\times h\times w\times c}$ before being fed into the denoising UNet. With this operation, the forward pass remains the same as the process without the view module. For cross-view information sharing, we reshape the feature map to $x \in \mathbb{R}^{(b\times t\times h\times w )\times m\times c}$ before the view module and then perform view attention, which refers to self-attention along the dimension $m$.


\begin{figure*}[t]
    \centering
    \includegraphics[width=0.9\textwidth]{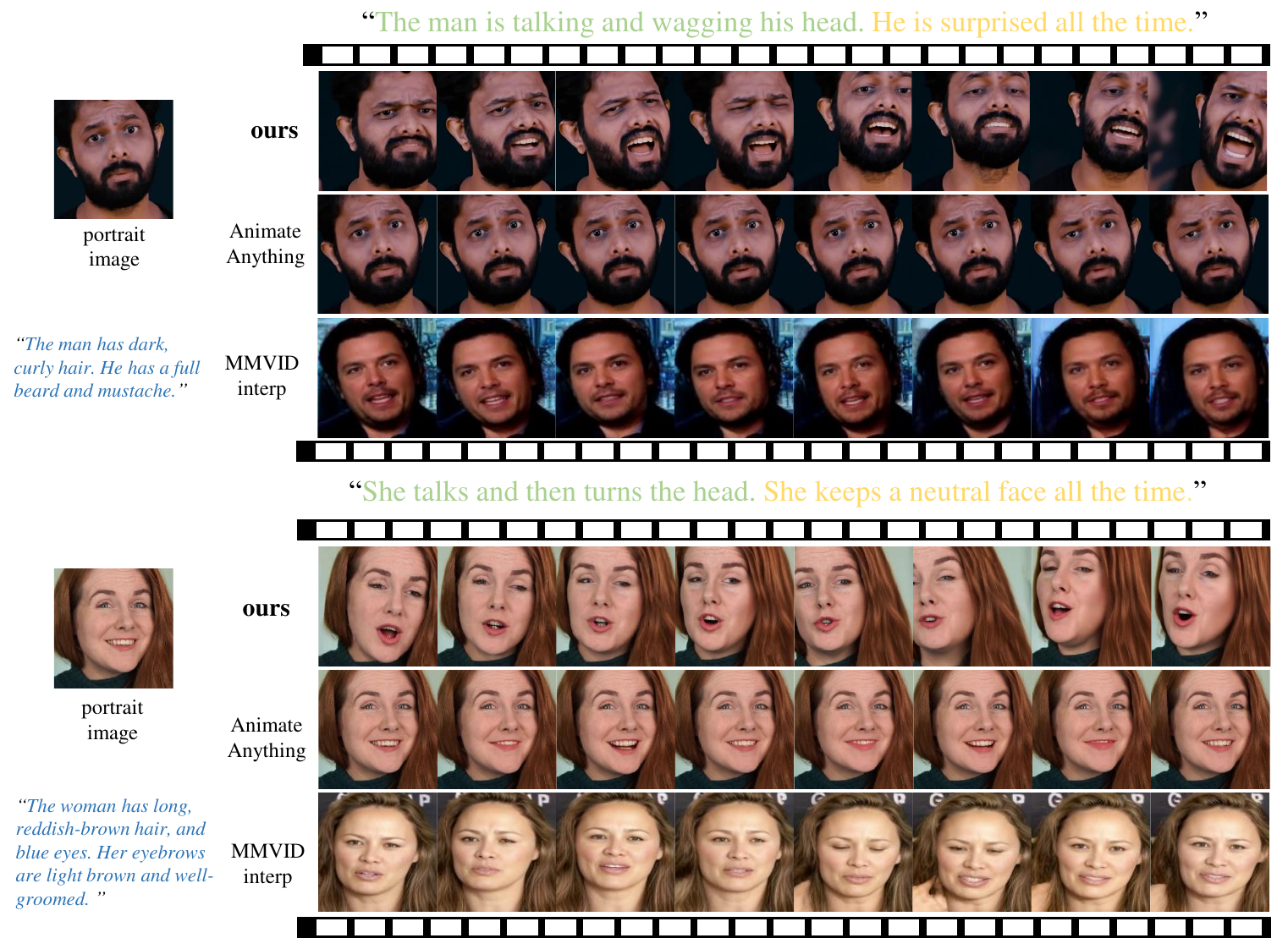}
    \caption{Qualitative comparison of text-guided portrait animation. \textcolor{green}{Motion descriptions} are highlighted in green, \textcolor{yellow}{emotion descriptions} in yellow, and \textcolor{blue}{appearance descriptions} (used only by MMVID-interp) in blue. Generated frames are shown sequentially from left to right. Additional examples and video results can be found in supplementary materials.}
    \label{qualitative}
\end{figure*}

\begin{figure*}[t]
    \centering
    \includegraphics[width=0.9\textwidth]{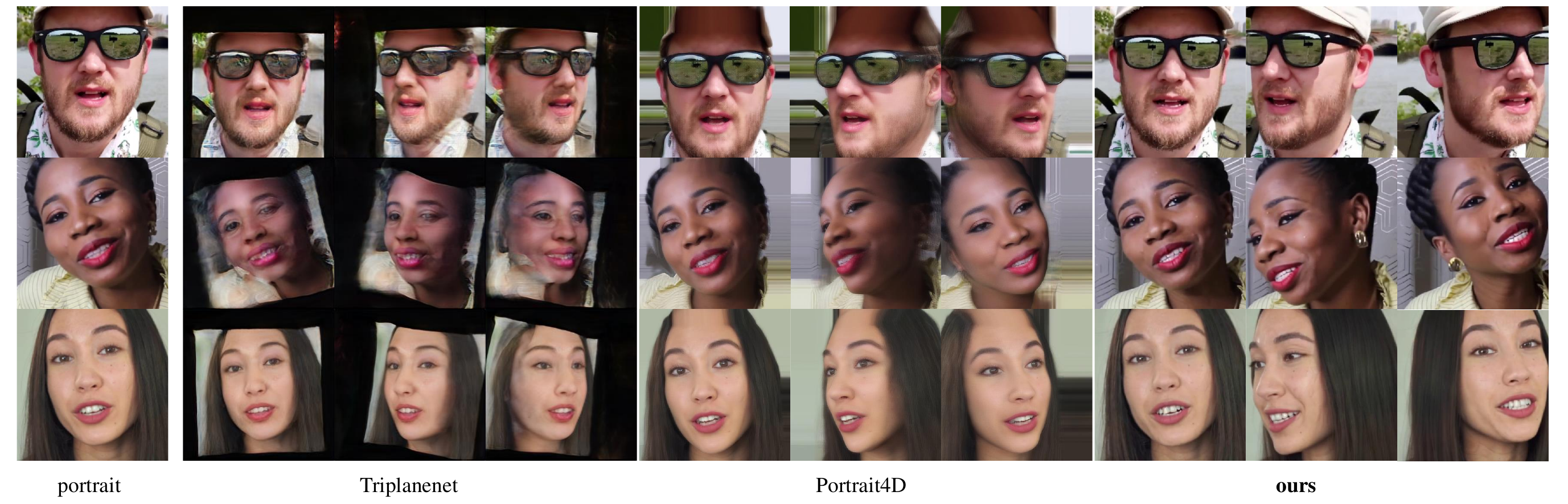}
    \caption{The qualitative comparison of multi-view consistency. We present results from $0^\circ$, $-30^\circ$ and $30^\circ$ perspectives.}
    \label{fig:comparison_multiview}
\end{figure*}

\begin{table*}[h]
  \begin{center}
    \begin{tabular}{l|cccc cc ccc}
    \toprule
      \textbf{Method} & LIQE$\uparrow$ & FID$\downarrow$ & FVD$\downarrow$ & CLIPSIM$\uparrow$ & VideoClip$\uparrow$ & Variability$\uparrow$ & MC$\uparrow$ &EC$\uparrow$ & VS$\uparrow$  \\
      \midrule
      AnimateAnything &4.024&34.9&\textbf{283.0}&0.171&0.564&0.095&1.33&1.23&1.41 \\

        MMVID-interp 
        &1.541&217.4&1548.9&0.175&0.517&0.092&1.67&1.56&2.33\\

        MVPortrait &\textbf{4.760}&\textbf{28.6}&570.0&$\textbf{0.183}$&$\textbf{0.595}$ &0.110&$\textbf{2.57}$&$\textbf{2.29}$&$\textbf{2.48}$\\
    
        \midrule
        No smoothing &3.849&108.4&1213.5&0.169&0.586&$\textbf{0.243}$&1.88&1.22&2.13\\
        
        Larger network &4.723&102.5&1572.9&0.174&0.548&0.064&1.44&1.75&1.38\\
        
        Joint training &4.409&76.8&824.7&0.175&0.559&0.068&1.50&1.44&1.56\\
        
        \bottomrule
    \end{tabular}
  \end{center}
  \vspace{-3mm}
  \caption{The quantitative comparison of text-guided portrait animation on the CelebV-Text test set.}
  \label{table:t2v}
\end{table*}

\begin{table*}[t!]
\centering
{
\begin{tabular}{lcccc|ccc}
\toprule
\multirow{2}{*}{Method}  & \multicolumn{4}{c}{\textbf{Self Reenactment}} & \multicolumn{3}{c}{\textbf{Cross Reenactment}} \\ \cmidrule(lr){2-5} \cmidrule(lr){6-8}
 & L1\ $\downarrow$  & SSIM\ $\uparrow$  & FID$\downarrow$ & FVD\ $\downarrow$ & ID\ $\uparrow$ & LIQE$\uparrow$ & FLAME-L1\ $\downarrow$ \\
\midrule

FollowYourEmoji \cite{ma2024follow}  & 92.346 & 0.763 & 68.5393 & 490.9643 & 0.888 & 4.0301 & 0.213 \\

LivePortrait \cite{guo2024liveportrait}  & \textbf{62.030} & 0.778 &  49.6557 & \textbf{268.7956} & 0.858 &  2.9312 & 0.213 \\

MVPortrait  & 74.875 & \textbf{0.780} & \textbf{48.4632}  & 409.3665 & \textbf{0.895} & \textbf{4.0605} &\textbf{0.196} \\ 
\bottomrule
\end{tabular}
}

\caption{ The quantitative comparison of video-driven portrait animation on the CelebV-Text test set.}
\label{tab:video-diven}
\end{table*}

\begin{table}[h]
  \begin{center}
    \small
    \setlength{\tabcolsep}{2pt}
    \begin{tabular}{l|ccc}
    \toprule
          \textbf{Method} & FID$\downarrow$ & FVD$\downarrow$  & SyncNet\cite{Chung16sync}$\downarrow$ \\
      \midrule

    SadTalker \cite{zhang2023sadtalker}
        &133.6572&644.2238& \textbf{13.5674}\\
        
      AniPortrait \cite{wei2024aniportrait} &\textbf{51.4862}&387.8381&14.4973 \\

        EchoMimic \cite{chen2024echomimic}
        &113.7123&596.6174&13.7097\\
        
        MVPortrait &56.1997& \textbf{333.1806}&14.6468\\
        
        \bottomrule
    \end{tabular}
  \end{center}
  \vspace{-3mm}
  \caption{The quantitative comparison of audio-driven portrait animation on the HDTF \cite{zhang2021hdtf} and CelebV-Text datasets.}
  \label{table:audio-driven}
\end{table}

\section{Experiment}
\subsection{Implementations} 
In the Text2FLAME stage, we utilize an encoder-transformer architecture for both MotionDM and EmotionDM, configured with a single layer and a latent dimension of 64. The velocity loss weight $\lambda_{\mathrm{vel}}$ is set to 0.5. For MotionDM, we adopt the 6D pose representation \cite{zhou2019continuity}, where $f_{\rm pose} \in \mathbb{R}^{12}$ represents the 6D pose of the head and jaw. We apply a sliding window of size 3 to smooth the pose and expression data. The noise schedule follows a cosine pattern, and diffusion is set with 1000 steps, paired with a learning rate of 1e-4. Both MotionDM and EmotionDM undergo 600k training steps, taking approximately 16 hours with a batch size of 64 on an A100 GPU.
In the FLAME2Video stage, we adopt a three-phase training strategy. The initial phase focuses on training the FLAME Encoder, Reference UNet, and the 2D components of the Denoising Unet, while excluding the motion module and view module. The motion module and view module are trained independently in Phase 2 and Phase 3, during which all other components are frozen. All images are resized to a resolution of $512 \times 512$, and the view number is set to 4. The Adam optimizer is used, with a constant learning rate of 1e-5. These three phases span approximately 3 days, 1 day, and 1 day respectively, on 8 A100 GPUs. 

In terms of speed, our pipeline averages 0.04 s/frame for Text2FLAME, with 0.021 seconds for motion and 0.019 seconds for emotion, 0.26 s/frame for FLAME rendering, and 0.45 s/frame for FLAME2Video. The Text2FLAME and FLAME2Video require 2GB and 14GB of VRAM, respectively.

\subsection{Dataset}
To learn the mapping from text to FLAME pose and expression, we utilize the CelebV-Text dataset \cite{yu2022celebvtext} to construct our CelebV-TF dataset, which includes over 15k text-FLAME pairs. CelebV-Text is a large-scale, high-quality, and diverse facial text-video dataset, with each video clip paired with texts describing the motion and emotion. From each video clip, we modify and employ DECA \cite{deca} to extract the FLAME parameters for Text2FLAME training and obtain the FLAME rendering images for FLAME2Video training. DECA is selected for its high efficiency, completing both estimation and rendering in approximately 0.3 s/frame. For multi-view animations training in the FLAME2Video stage, we collect a subset of RenderMe-360 Dataset \cite{pan2024renderme}, which consists of 21 performers. Aligning the goal of our model, we intentionally choose video clips of 17 different motions while the selected 27 camera perspectives fulfill the need for frontal and side views of the face. We also employ DECA to extract the FLAME parameters and corresponding rendering images.



\subsection{Baseline and Metric}
It is important to note that the primary functions of our framework are text-guided portrait animation and multi-view video generation. Therefore, our evaluation can be divided into two parts: the effectiveness of text-guided single-view portrait animation (at the video level) and the consistency of multi-view generation (at the image level).

For text-guided single-view portrait animation, we compare our approach with AnimateAnything \cite{dai2023animateanything} and MMVID-interp \cite{yu2022celebvtext}, both fine-tuned on the CelebV-Text dataset. Following \cite{yu2022celebvtext}, we employ FID \cite{FID} for per-frame quality, FVD \cite{unterthiner2018fvd} for temporal coherence, and CLIPSIM \cite{wu2021clipsim} for text-video relevance. We also include LIQE \cite{zhang2023liqe} for image quality assessment and VideoCLIP \cite{askvideos2024videoclip} to evaluate video-level text-video alignment. To evaluate the expressiveness of videos, we designed a metric called ``Variability" to calculate the degree of head movement changes. In addition, a user study was conducted to assess animation quality across three subjective criteria: Motion Consistency (MC), Emotion Consistency (EC), and Vividness Score (VS). Ratings were assigned on a 3-point scale: 1 (poor), 2 (average), and 3 (good). Participants were tasked with rating each video. As for multi-view generation, we compare with Triplanenet \cite{bhattarai2024triplanenet} and  Portrait4D-v2 \cite{deng2024portrait4d}, both of which are for face novel view synthesis. We utilize LPIPS \cite{lpips} and SSIM \cite{SSIM} for evaluation of image quality, ID \cite{deng2019arcface} for multi-view identity consistency. Refer to the supplementary materials for the implementation of all metrics.

Since our framework supports both speech-driven and video-driven portrait animation, we compare it with methods from each of these categories. For the FLAME-L1 metric, we estimate the FLAME parameters for both the driving video and the generated video, and then calculate the difference between them. The quantitative comparison results can be found in \cref{tab:video-diven} and \cref{table:audio-driven}. Refer to the supplementary materials for a comprehensive visual comparison.

\subsection{Qualitative Results}
As observed in \cref{qualitative}, with the prompt “The man is talking and wagging his head. He is surprised all the time,” AnimateAnything produces a nearly static result with minimal mouth movement and no head wagging, missing the surprise expression entirely. MMVID-interp captures some mouth movement and slight head motion, but lacks a consistent surprise expression across frames. Besides, the videos exhibit visible distortion and unnatural transitions. In contrast, our approach effectively generates both talking and head-wagging motions, with a clear, continuous expression of surprise. Overall, our approach generates highly dynamic and expressive results, with head and mouth movements and expressions that are closely aligned with the text prompts.


We present the comparison of multi-view synthesis in Fig. \ref{fig:comparison_multiview}. We can see that Triplanenet often produces severe artifacts and exhibits poor multi-view consistency. While Portrait4D-v2 obtains consistent multi-view portraits, it suffers from blurriness around the forehead. In contrast, our method ensures both image quality and consistent multi-view representation.

\subsection{Quantitative Results}

As shown in \cref{table:t2v}, our method outperforms both AnimateAnything and MMVID-interp on CLIPSIM and VideoClip metrics, improving the alignment between generated videos and text prompts. Besides, we surpass the others on the Variability metric, indicating much more dynamic animations. Additionally, our approach leads in subjective evaluation, with notable improvements in MC, EC, and VS, suggesting that users find our videos with better alignment in motion and emotion, as well as a more vivid and expressive representation. Although our method shows lower performance than AnimateAnything in terms of FVD, this is due to the larger head motion amplitude in our videos, which causes some misalignment with the original dataset. However, our approach excels in expressiveness, capturing dynamic head movements and emotions, whereas AnimateAnything produces nearly static results. 

Furthermore, as demonstrated in \cref{tab:video-diven} and \cref{table:audio-driven}, our unified animation framework demonstrates strong competitiveness in both video-driven and audio-driven scenarios.

\begin{table}[h]
  \begin{center}
    \label{quantitative_GSO}
    \small
    \setlength{\tabcolsep}{3.2pt}
    \begin{tabular}{l|ccc}
    \toprule
      \textbf{Method} & LPIPS$\downarrow$ & SSIM$\uparrow$  & ID$\uparrow$\\
      
      \midrule
      Triplanenet
      &\textbf{0.0936}&0.5974&0.7710\\

        Portrait4D-v2
       &0.4468&0.5278&0.8006\\
        
        MVPortrait &0.2101&$\textbf{0.6224}$&$\textbf{0.8409}$\\
        \midrule
        w/o view attention &0.2268&0.6058&0.7149 \\
     \bottomrule
    \end{tabular}
  \end{center}
  \vspace{-2mm}
   \caption{The comparison of multi-view portrait synthesis.}
   \label{table:quantitative_mv}
\end{table}

\cref{table:quantitative_mv} shows the numerical comparison of multi-view synthesis. We argue that LPIPS favors images with black backgrounds, which is why Triplanenet achieves unusual results. Our method is superior to others in SSIM by a large margin, demonstrating the most compelling image quality. We achieve an ID score of 0.8409, which indicates that our method provides excellent multi-view identity consistency. This proves that FLAME encapsulates sufficient facial information and maintains consistency across multiple viewpoints, facilitating accurate and coherent reproduction of the character's appearance. More comparisons of multi-view synthesis are provided in the supplementary materials.


\subsection{Ablation Study}
\begin{figure}[h]
    \centering
    \includegraphics[width=\linewidth]{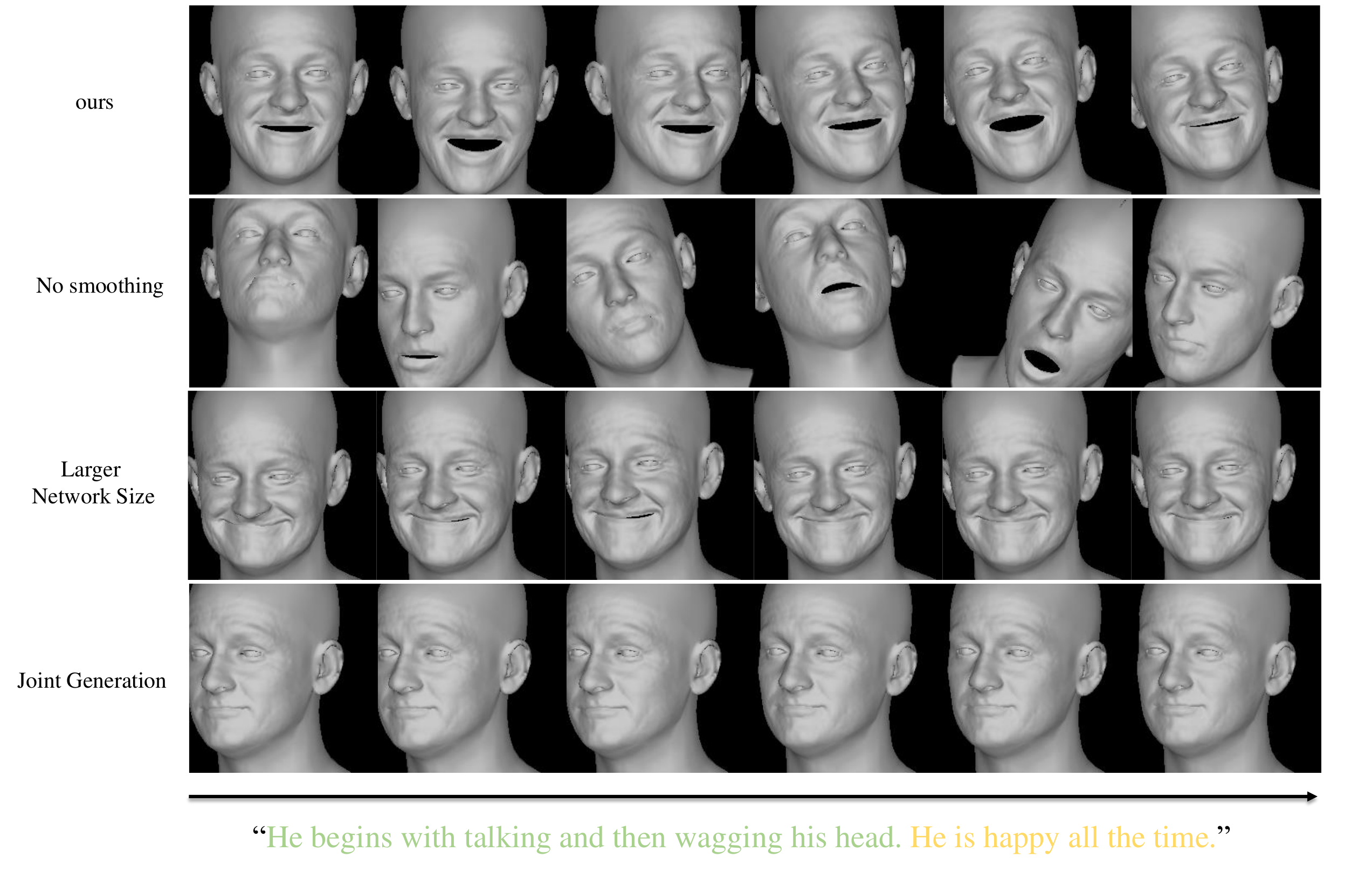}
    \caption{The ablation of the Text2FLAME stage.}
    \label{fig: t2f_ablation}
\end{figure}

\textbf{Text2FLAME.} We ablate the Text2FLAME stage using three variants: \textit{No Smoothing}, and \textit{Larger Network Size}, \textit{Joint Generation}. The No Smoothing variant involves training on raw data without sliding average smoothing. For the Larger Network Size variant, we expand the network to 4 layers with a latent dimension of 128. For the Joint Generation variant, we train a single model to generate both motion and emotion simultaneously, as opposed to our baseline approach of separate training. 
As illustrated in \cref{fig: t2f_ablation}, our results align closely with the text prompt, demonstrating natural and expressive mouth and head movements. In contrast, the No Smoothing variant produces mismatched expressions and severe head jittery, thus leading to a higher Variability score compared to MVPortrait.  The Larger Network Size variant successfully generates correct expressions, but the head movements are nearly absent. Finally, the Joint Generation variant shows both incorrect expressions and static motion, further emphasizing the benefits of our separate training in MotionDM and EmotionDM.

\begin{figure}[h]
    \centering
    \includegraphics[width=\linewidth]{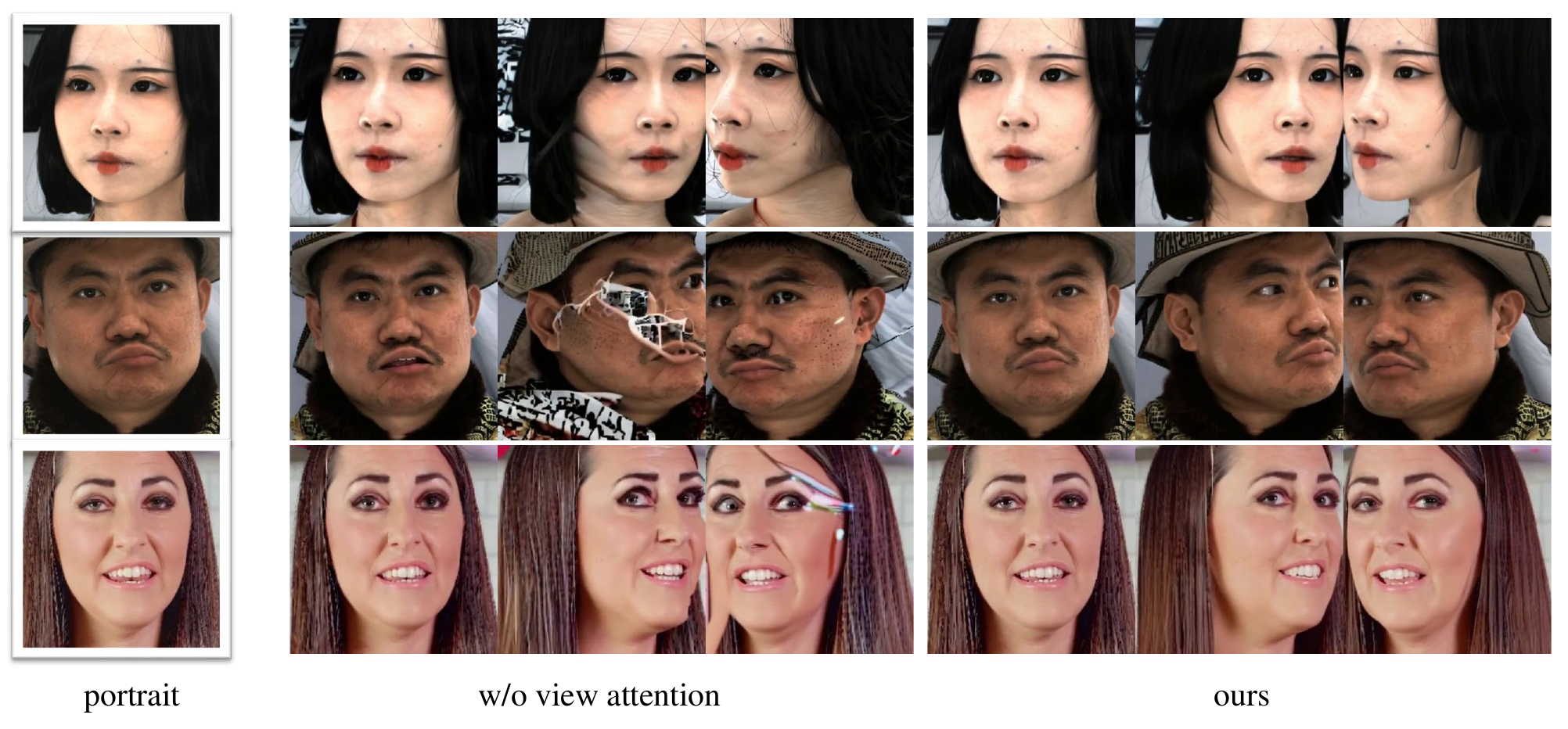}
    \caption{The ablation of the view attention.}
    \label{fig: f2v_ablation}

\end{figure}

\textbf{FLAME2Video.} In this stage, we evaluate the effectiveness of view attention in ensuring multi-view consistency. As shown in \cref{fig: f2v_ablation}, removing the view module results in severe artifacts on the face in the new viewpoints, and the character's appearance deviates from the original portrait. This highlights the importance of the view module for achieving realistic and view-consistent animations. 

\section{Conclusion}
In conclusion, MVPortrait addresses key challenges in portrait animation by enhancing control over motion, emotion, and multi-view consistency. By employing the FLAME model as an intermediate representation, our two-stage framework effectively integrates these aspects to produce more expressive and consistent animations.
Moreover, MVPortrait’s unique ability to integrate text, audio, and video as driving signals offers unprecedented flexibility and control, making it a significant advancement in the field of portrait animation, particularly text-driven animation.


However, MVPortrait is limited by the accuracy of text annotations in video datasets and the uneven distribution of facial movements and emotions. Besides, despite FLAME's proficiency in preserving identity, it has limitations in capturing micro-expressions, which remains a direction for future exploration. In the future, we will design a text-driven framework with finer-grained control over expression, while also improving its generalization ability.

\section{Acknowledgements}
This work was partially supported by the Shenzhen Key Laboratory of next generation interactive media innovative technology (No.ZDSYS20210623092001004).
\clearpage
{
    \small
    \bibliographystyle{ieeenat_fullname}
    \bibliography{main}
}

\setcounter{page}{1}
\maketitlesupplementary


\section{Training}
The loss for FLAME2Video stage is formulated as 
\begin{equation}
\mathbb{E}_{\epsilon\thicksim\mathcal{N}(\mathbf{0},\mathbf{I}),\mathbf{x}_t,\mathbf{c},t}[\|\epsilon-\epsilon_\theta(\mathbf{x}_t;\mathbf{c},t)\|_2^2]
\end{equation}
where $\mathbf{c}$ is image embedding encoded by CLIP and $\mathbf{x}_t$ is derived by adding $t$-step noises to $\mathbf{x}_0$, and $\epsilon$ and $\epsilon_\theta$ respectively represent the ground truth and predicted noise by Denoising UNet.

In the Text2FLAME stage, the conditional learning strategy we use is a classifier-free guidance. We set the text condition to null with a probability of 10\%. Following \cite{tevet2023human},  we set a maximum frame limit during the 
training, with padding applied to shorter videos. The diffusion sampling step is set to 1000. In the FLAME2Video stage, the diffusion sampling step is set to 25. We utilize the pre-trained UNet weights from runwayml/stable-diffusion-v1-5.

\section{Metrics}
In this section, we explain the calculation processes for the metrics not detailed in the main text.

\textbf{Variability}. We devise a metric to evaluate the amplitude of head movements to assess the expressiveness of portrait videos. Specifically, DECA \cite{deca} is employed to estimate FLAME parameters, through which the yaw, pitch, and roll angles are extracted from the head pose. The mean variability of motion in the generated videos is quantified by calculating the temporal differences between consecutive frames for these pose angles.

\textbf{FLAME-L1}. To quantify the differences in pose and expression between the generated and driving videos during cross-reenactment, we utilize DECA \cite{deca} to estimate FLAME parameters, including pose and expression, for both the generated and driving videos. The FLAME-L1 metric is then computed by calculating the L1 difference between the corresponding parameters.

\begin{table*}[htb]
  \centering
  \begin{tabular}{l|ccccccccc}
    \toprule
    \textbf{Method} & LIQE$\uparrow$ & FID$\downarrow$ & FVD$\downarrow$ & CLIPSIM$\uparrow$ & VideoClip$\uparrow$ & Variability$\uparrow$ & MC$\uparrow$ & EC$\uparrow$ & VS$\uparrow$ \\
    \midrule
    DynamiCrafter-ft & 4.306 & 80.9 & \textbf{552.8} & 0.172 & 0.557 & 0.074 & 1.70 & 1.22 & 1.78 \\
    CogVideoX-ft & 4.046 & 171.7 & 962.3& 0.179 & 0.589 & 0.119 & 2.37 & 2.22 & 2.25 \\
    MVPortrait & \textbf{4.760} & \textbf{28.6} & 570.0 & \textbf{0.183} & \textbf{0.595} & 0.110 & \textbf{2.57} & \textbf{2.29} & \textbf{2.48} \\
    \midrule
    No smoothing &3.849&108.4&1213.5&0.169&0.586&$\textbf{0.243}$&1.88&1.22&2.13\\
    Window size: 5 & 4.590 & 67.0 & 948.7 & 0.180 & 0.593 & 0.104 & 2.25 & 2.10 & 2.11 \\
    Joint training & 4.409 & 76.8 & 824.7 & 0.175 & 0.559 & 0.068 & 1.50 & 1.44 & 1.56 \\
    OOD cases & 4.245 & - & - & 0.175 & 0.561 & 0.111 & 1.67 & 2.11 & 1.89 \\
    \bottomrule
  \end{tabular}
  \caption{Comparison of text-guided animation on the CelebV-Text test set.}
  \label{table:reb-t2v}
\end{table*}

\begin{table*}[htb]
  \begin{center}
    \begin{tabular}{l|cccc}
    \toprule
      \textbf{Method} & LPIPS$\downarrow$ & SSIM$\uparrow$ & L1$\downarrow$ & ID$\uparrow$ \\
      
      \midrule
      Triplanenet-ft
      &0.3752&0.5111& 0.2200 &0.8803\\

        Portrait4D-v2-ft
       & 0.3725& 0.5150 & 0.1957 & 0.8342\\
        
        MVPortrait (view number: 4)&\textbf{0.2445}&$\textbf{0.5512}$& \textbf{0.1735} &$\textbf{0.8891}$\\
        \midrule
        view number: 2 & 0.3206 & 0.5370 & 0.1971 & 0.8224\\
        w/o view attention (view number: 1) &0.3959&0.4624&0.2746&0.7826\\
     \bottomrule
    \end{tabular}
  \end{center}
   \caption{Comparison of multi-view portrait synthesis on the RenderMe-360 test set.}
   \label{table:reb-quantitative_mv}
\end{table*}

\section{Quantitative Comparison}

\subsection{Comparison with more baselines}
We compare with text-guided video generative models like CogVideoX\cite{yang2024cogvideox} and DynamiCrafter\cite{xing2024dynamicrafter} to benchmark our model's performance against current leading methods. We fine-tune them on CelebV-Text, with test results in \cref{table:reb-t2v}.

\subsection{Out-of-Distribution Performance}
\begin{figure}[h]
    \centering
    \includegraphics[width=\linewidth]{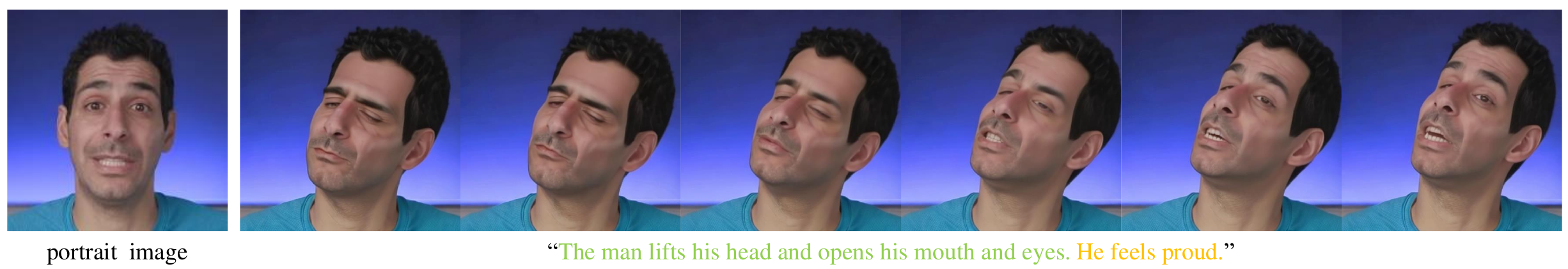}
    \caption{The example for demonstrating out-of-distribution performance.}
    \label{fig: reb}
\end{figure}

We generate 10 out-of-distribution (OOD) text descriptions using GPT-4, and present the quantitative results for these cases in \cref{table:reb-t2v}, where a performance drop is observed. However, the example in \cref{fig: reb} illustrates that our model shows some generalization capability. 

\subsection{Multi-view Comparison}
For fairness, we fine-tune Triplanenet \cite{bhattarai2024triplanenet} and Portrait4D-v2 \cite{deng2024portrait4d} on RenderMe-360 \cite{pan2024renderme} training set, and present the evaluation results on the RenderMe-360 test set in \cref{table:reb-quantitative_mv}. Multi-view accuracy is measured as the average LPIPS, SSIM, and L1 differences between the predictions and ground truth across all viewpoints.

\section{Qualitative Comparison}
We present additional visual comparisons to provide readers with a clearer view of the differences between various methods. Since FLAME \cite{FLAME} acts as an intermediate representation, our framework becomes the first to support text, video, and audio-driven portrait animations. In the following, we present qualitative comparisons of our method with others under different driving signals.

\subsection{Text-driven Animation}
\begin{figure*}[htb]
    \centering
    \includegraphics[width=\textwidth]{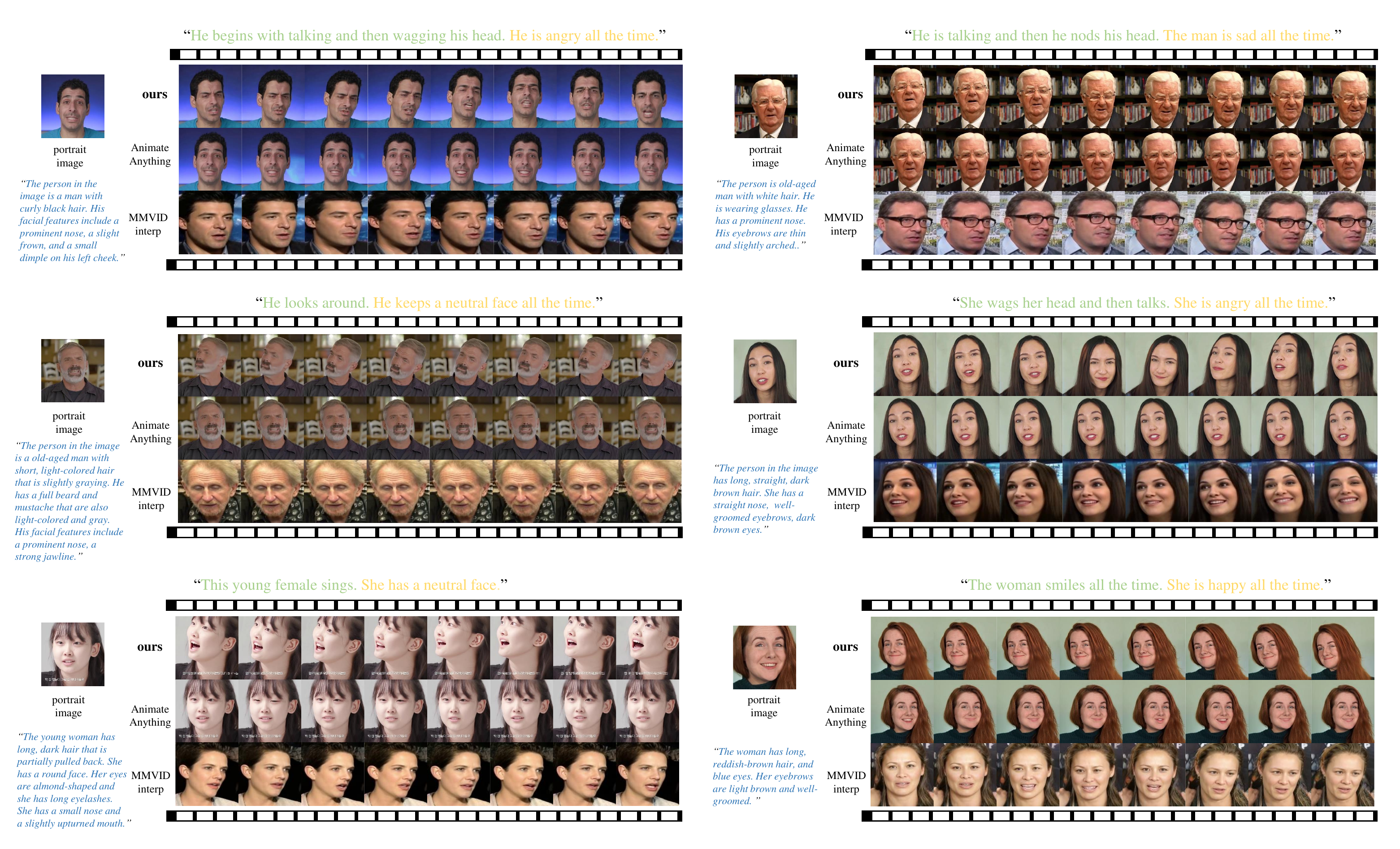}
    \caption{The qualitative comparison of text-guided portrait animation. The motion descriptions are highlighted in \textcolor{green}{green}, while the emotion descriptions are marked in \textcolor{yellow}{yellow}. The \textcolor{blue}{blue} text represents the appearance descriptions, which is only used by MMVID-interp. The generated video frames are displayed sequentially from left to right. }
    \label{suppl_text}
\end{figure*}

For text-driven portrait animation, we compare our method with AnimateAnything \cite{dai2023animateanything} and MMVID-interp \cite{yu2022celebvtext}, both of which were fine-tuned on the CelebV-Text dataset to ensure a fair comparison. As end-to-end generation models, AnimateAnything and MMVID-interp achieve text-driven portrait video generation by learning an implicit mapping between text and video. However, both methods exhibit weaker controllability compared to our approach, which leverages FLAME for explicit control. This advantage is primarily due to our use of a text-guided diffusion model to generate the corresponding head poses and facial expressions. As shown in \cref{suppl_text}, our method surpasses the other two methods in terms of motion and emotion consistency with the text description, and demonstrates superior vividness. Furthermore, videos are provided below for readers to review.

\subsection{Video-driven Animation}
Given a driving video, we first use the FLAME estimation method, DECA \cite{deca} in our experiment to estimate the corresponding FLAME sequence and obtain the renderings. Next, we use these renderings to generate the animated video. Our experiments reveal that videos generated by FollowYourEmoji exhibit significant flickering artifacts, as evident in the video. While LivePortrait demonstrates strong driving performance, the pose and expression in its generated videos often fail to align with those in the driving video. In contrast, our method produces videos with superior robustness and controllability. 

\begin{figure*}[htb]
    \centering
    \includegraphics[width=\textwidth]{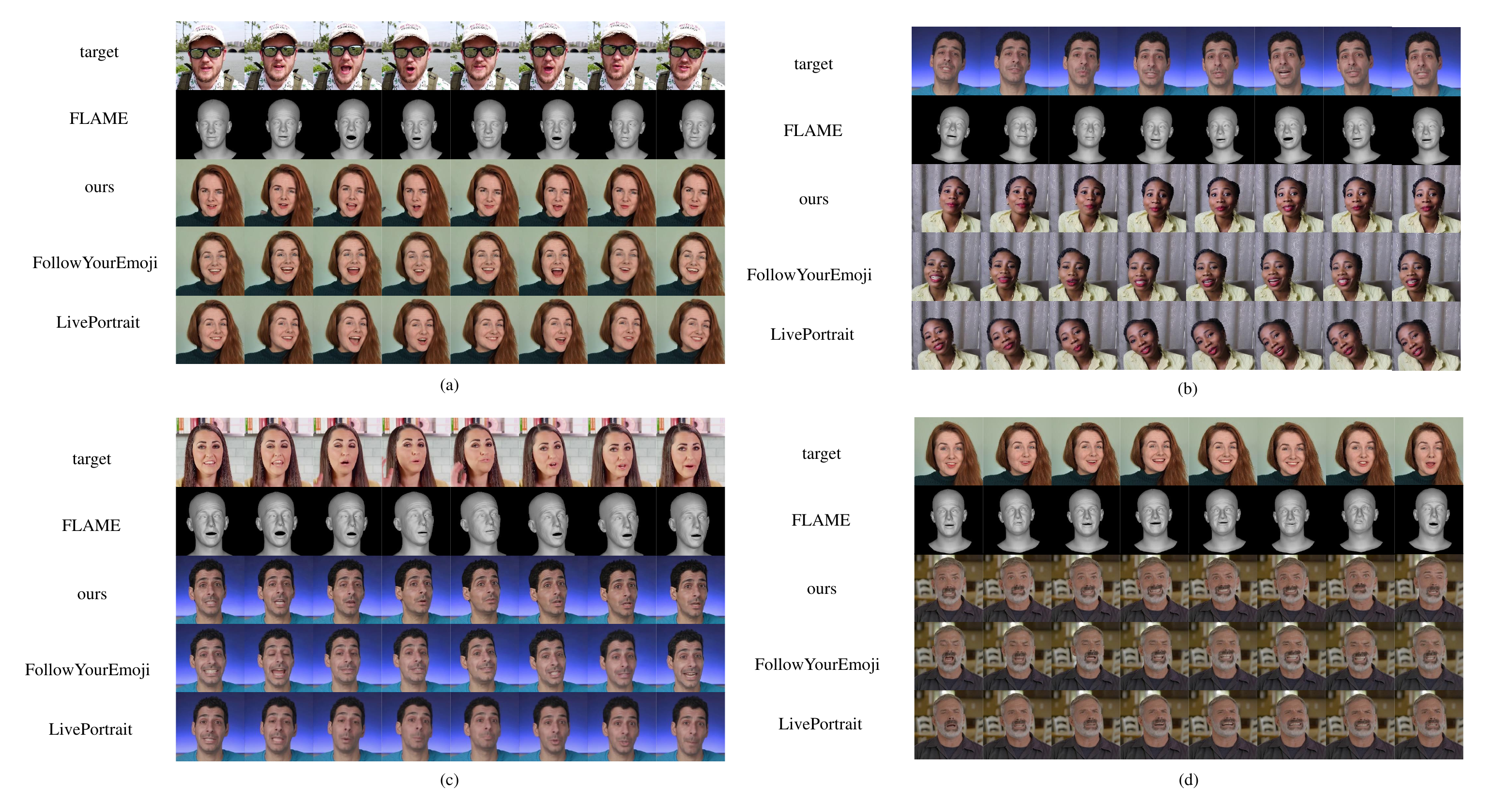}
    \caption{The qualitative comparison of video-driven portrait animation.  The generated video frames are displayed sequentially from left to right. }
    \label{suppl_video}
\end{figure*}

Visual comparisons are provided in \cref{suppl_video}. In each subplot, the first row shows the driving video, and the second row shows the FLAME sequence, which is constructed from the FLAME pose and expression parameter sequences estimated from the driving video, along with the facial shape parameters and facial detail vectors estimated from the reference image. Thus, the FLAME sequence here represents both the head movements and facial expressions exhibited in the driving video, as well as the facial shape and details of the reference portrait. These FLAME sequences can serve as a reference for evaluating the effectiveness of the driving algorithm. It is clear from subplots (a) and (b) that the head pose in the results generated by LivePortrait is significantly inconsistent with the driving video. Additionally, in subplot (c), severe blurring is observed in the video generated by LivePortrait, which may be due to the presence of hands in the driving video, a scenario that LivePortrait struggles to handle robustly. FollowYourEmoji also struggles to handle head pose and tends to produce larger mouth movements compared to the driving video, as shown in all subplots. We encourage readers to watch the videos in the supplementary materials to gain a more intuitive understanding of the differences between the methods.

\subsection{Audio-driven Animation}
\begin{figure*}[htb]
    \centering
    \includegraphics[width=\textwidth]{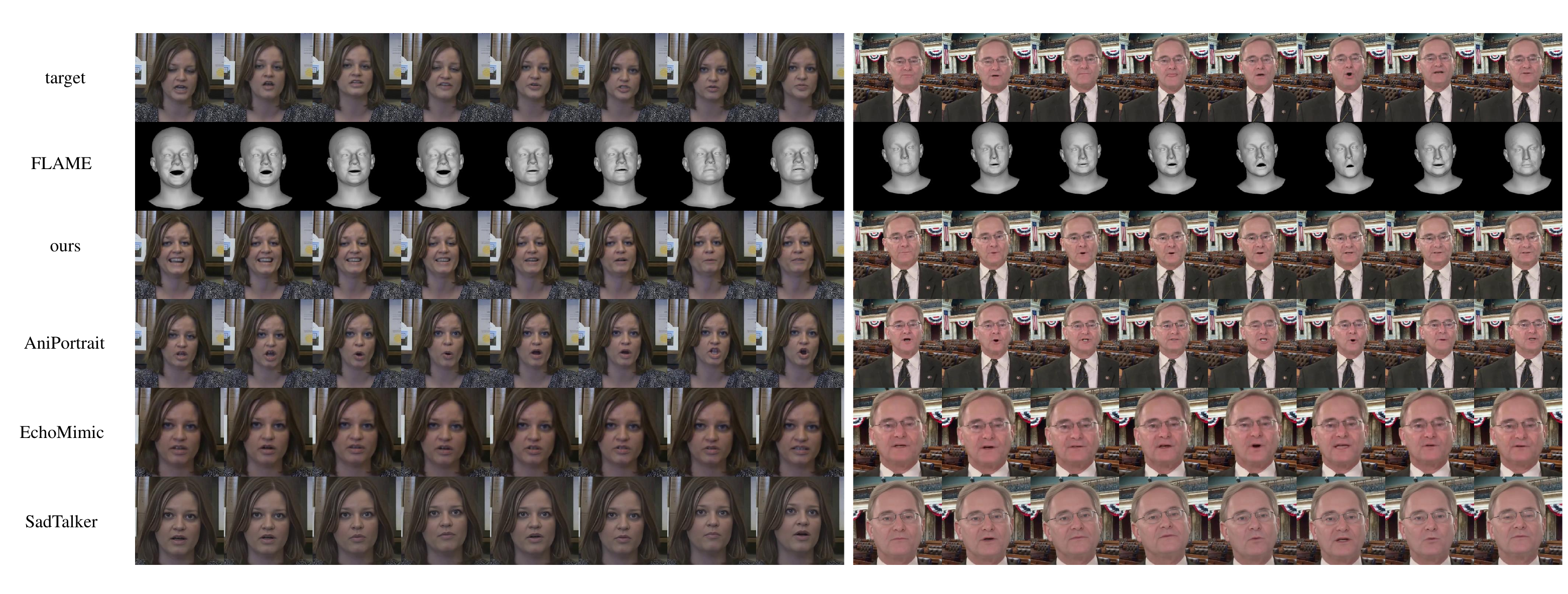}
    \caption{The qualitative comparison of audio-driven portrait animation.  The generated video frames are displayed sequentially from left to right. }
    \label{suppl_audio}
\end{figure*}

In our framework, audio-driven generation is also carried out in two stages. In the first stage, we use an audio-driven head generation method, TalkShow \cite{yi2023generating} in our experiment, to produce FLAME parameters. The FLAME parameters are used for generating FLAME images as guidance conditions. In the second stage, FLAME is used to create the animation. When showcasing MVPortrait's performance in audio-driven animation, we also present the FLAME generated by TalkShow, to better assess the synchronization between audio and video. Refer to \cref{suppl_audio} for comparison.

\begin{figure*}[htb]
    \centering
    \includegraphics[width=\textwidth]{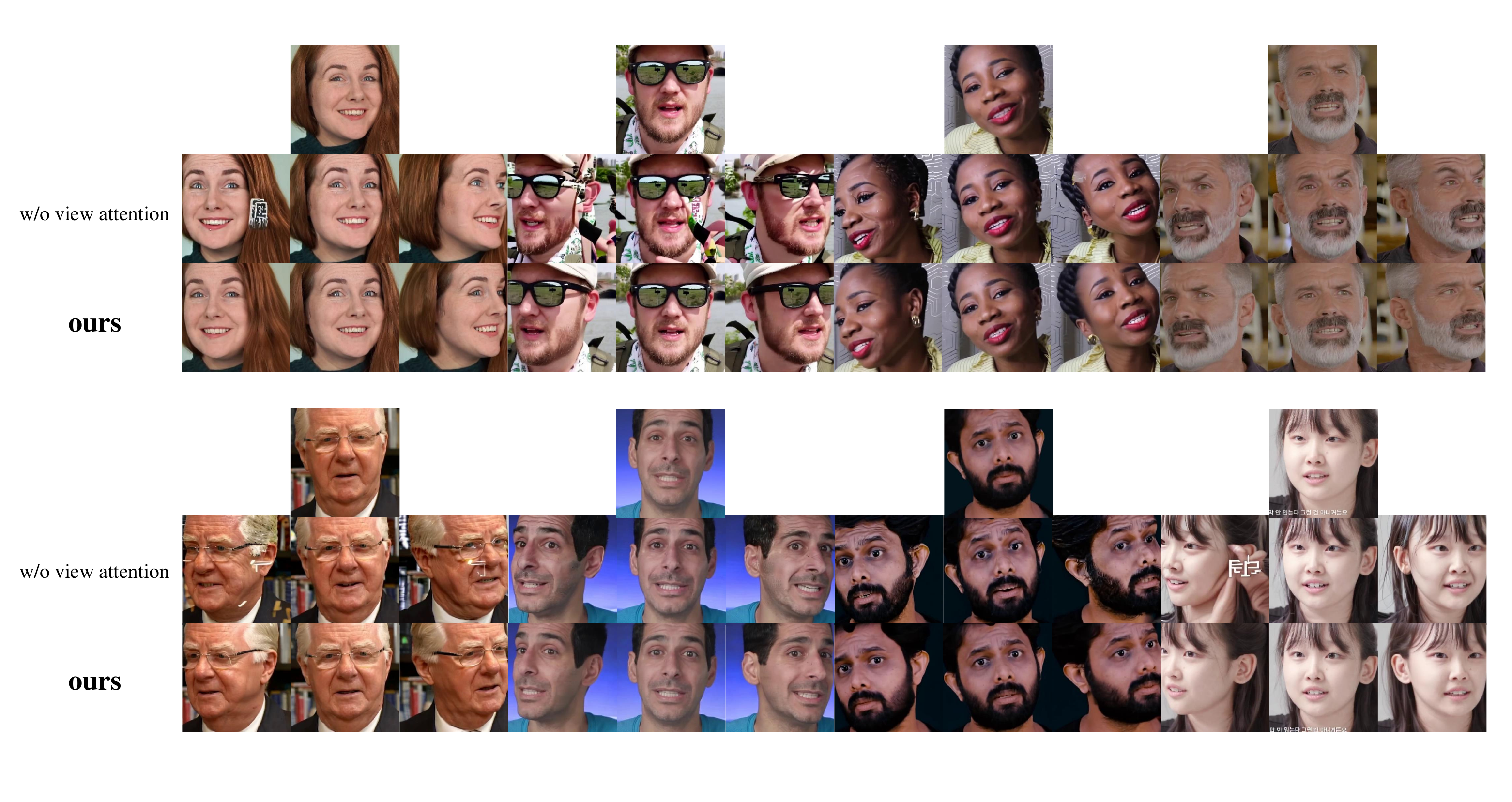}
    \caption{The qualitative ablation of view attention.}
    \label{suppl_multiview}
\end{figure*}

\section{Ablation}
\subsection{Text2FLAME}
We ablate the Text2FLAME stage using three variants: \textit{No Smoothing}, and \textit{Larger Network Size}, \textit{Joint Generation}. As mentioned in the main text, the No Smoothing variant causes mismatched expressions and head jitter, the Larger Network Size variant generates correct expressions but lacks head movement, and the Joint Generation variant shows incorrect expressions and static motion. We include animations of these variants in the supplementary materials. Please watch them for comparison. 

Additionally, for smoothing operations, we conduct an additional experiment with a sliding window size of 5. The quantitative results are shown in \cref{table:reb-t2v}. Note that the window size for MVPortrait is 3, while the window size for \textit{no smoothing} is 1. This ablation shows a window size of 3 balances stability and naturalness best.

\subsection{FLAME2Video}
In this part, we present additional multiview results to demonstrate the effectiveness of our view attention mechanism in \cref{suppl_multiview}. We train a 2-view model and 
report evaluation results in \cref{table:reb-quantitative_mv}, which shows that the performance improves as the number of views increases, up to the maximum of 4 supported on an 80GB A100 GPU.

\end{document}